\documentclass[11pt]{article}
\usepackage[utf8]{inputenc}
\usepackage{amsfonts}
\usepackage{amsmath}
\usepackage{amsthm}
\usepackage{color}
\usepackage{newtxtext}
\usepackage{graphicx}
\usepackage[margin=1in]{geometry}
\usepackage{mathtools}
\usepackage{booktabs}
\usepackage{authblk}
\usepackage{url}
\graphicspath{{figs/}}

\newtheorem{theorem}{Theorem}

\newtheorem{lemma}[theorem]{Lemma}
\newtheorem{definition}{Definition}

\title{DeepONet: Learning nonlinear operators for identifying differential equations based on the universal approximation theorem of operators}
\author[1]{Lu Lu}
\author[2]{Pengzhan Jin}
\author[1]{George Em  Karniadakis}
\affil[1]{Division of Applied Mathematics, Brown University, Providence, RI 02912, USA}
\affil[2]{LSEC, ICMSEC, Academy of Mathematics and Systems Science, Chinese Academy of Sciences, Beijing 100190, China}
\date{}

\begin{document}
\maketitle

\begin{abstract}
While it is widely known that neural networks are universal approximators of continuous functions, a less known and perhaps more powerful result is that a neural network with a single hidden layer can approximate accurately any nonlinear continuous operator \cite{chen1995universal}. This universal approximation theorem is suggestive of the potential application of neural networks in learning nonlinear operators from data. However, the theorem guarantees only a small approximation error for a sufficient large network, and does not consider the important optimization and generalization errors. To realize this theorem in practice, we propose deep operator networks (DeepONets) to learn operators accurately and efficiently from a relatively small dataset. A DeepONet consists of two sub-networks, one for encoding the input function at a fixed number of sensors $x_i, i=1,\dots,m$ (branch net), and another for encoding the locations for the output functions (trunk net).
We perform systematic simulations for identifying two types of operators, i.e., dynamic systems and partial differential equations, and demonstrate that DeepONet significantly reduces the generalization error compared to the fully-connected networks. We also derive theoretically the dependence of the approximation error in terms of the number of sensors (where the input function is defined) as well as the input function type, and we verify the theorem with computational results. More importantly, we observe high-order error convergence in our computational tests, namely polynomial rates (from half order to fourth order) and even exponential convergence with respect to the training dataset size.
\end{abstract}

\section{Introduction}

The universal approximation theorem states that neural networks can be used to approximate any continuous function to arbitrary accuracy if no constraint is placed on the width and depth of the hidden layers \cite{cybenko1989approximation,hornik1989multilayer}. However,
another approximation result, which is yet more surprising and has not been appreciated so far, states that a neural network with 
a single hidden layer can approximate accurately any nonlinear continuous {\em functional} (a mapping from a space of functions into the real numbers)~\cite{chen1993approximations,mhaskar1997neural,rossi2005functional} or (nonlinear) operator (a mapping from a space of functions into another space of functions)~\cite{chen1995universal,chen1995approximation}.

\begin{figure}[htbp]
\centering
\includegraphics{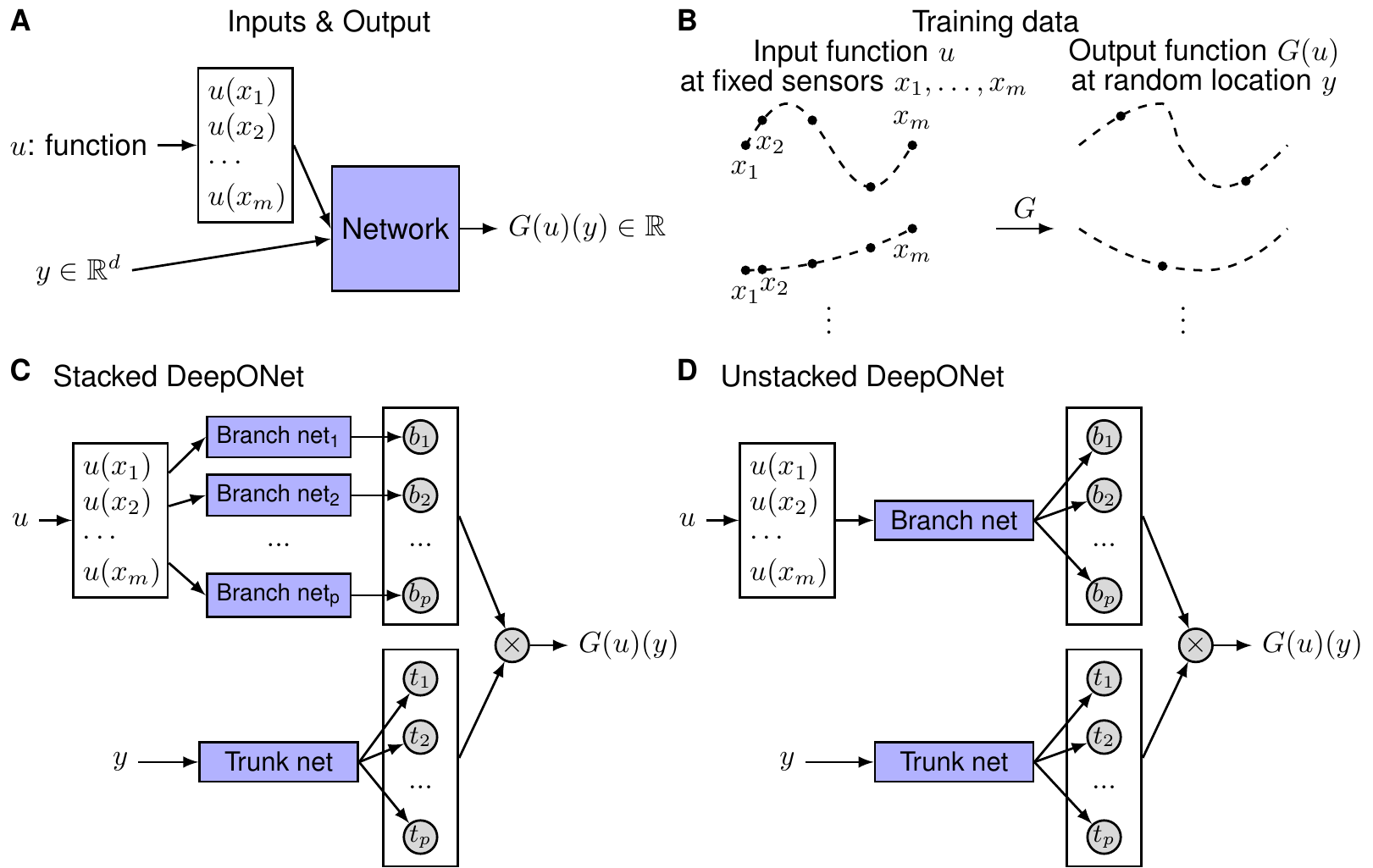}
\caption{\textbf{Illustrations of the problem setup and architectures of DeepONets.} (\textbf{A}) The network to learn an operator $G: u \mapsto G(u)$ takes two inputs $[u(x_1), u(x_2), \dots, u(x_m)]$ and $y$. (\textbf{B}) Illustration of the training data. For each input function $u$, we require that we have the same number of evaluations at the same scattered sensors $x_1, x_2, \dots, x_m$. However, we do not enforce any constraints on the number or locations for the evaluation of output functions. (\textbf{C}) The stacked DeepONet in Theorem~\ref{thm:main} has one trunk network and $p$ stacked branch networks. (\textbf{D}) The unstacked DeepONet has one trunk network and one branch network.}
\label{fig:problem}
\end{figure}

Before reviewing the approximation theorem for operators, we introduce some notation, which will be used through this paper. Let $G$ be an operator taking an input function $u$, and then $G(u)$ is the corresponding output function. For any point $y$ in the domain of $G(u)$,  the output $G(u)(y)$ is a real number. Hence, the network takes inputs composed of two parts: $u$ and $y$, and outputs $G(u)(y)$ (Fig. \ref{fig:problem}A). Although our goal is to learn operators, which take a function as the input, we have to represent the input functions discretely, so that network approximations can be applied. A straightforward and simple way, in practice, is to employ the function values at sufficient but finite many locations $\{x_1, x_2, \dots, x_m\}$; we call these locations as ``sensors'' (Fig. \ref{fig:problem}A). Next, we state the following theorem due to Chen \& Chen \cite{chen1995universal}, see appendix for more details.

\begin{theorem}[\textbf{Universal Approximation Theorem for Operator}]
\label{thm:main}
Suppose that $\sigma$ is a continuous non-polynomial function, $X$ is a Banach Space, $K_1 \subset X$, $K_2 \subset \mathbb{R}^d$ are two compact sets in $X$ and $\mathbb{R}^d$, respectively, $V$ is a compact set in $C(K_1)$, $G$ is a nonlinear continuous operator, which maps $V$ into $C(K_2)$. Then for any $\epsilon>0$, there are positive integers $n$, $p$, $m$, constants $c_i^k, \xi_{ij}^k, \theta_i^k, \zeta_k \in \mathbb{R}$, $w_k \in \mathbb{R}^d$, $x_j \in K_1$, $i=1,\dots,n$, $k=1,\dots,p$, $j=1,\dots,m$, such that
\begin{equation}\label{eq:thm}
\left|G(u)(y) - \sum_{k=1}^p
\underbrace{\sum_{i=1}^n c_i^k \sigma\left(\sum_{j=1}^m \xi_{ij}^ku(x_j)+\theta_i^k\right)}_{branch}
\underbrace{\sigma(w_k \cdot y+\zeta_k)}_{trunk}
\right|<\epsilon  
\end{equation}
holds for all $u \in V$ and $y \in K_2$.
\end{theorem}

This approximation theorem indicates the potential application of neural networks to learn nonlinear operators from data, i.e., similar to what the deep learning community is currently doing, that is learning functions from data. However, this theorem does not inform us how to learn operators effectively. Considering the classical image classification task as an example, the universal approximation theorem of neural networks for functions \cite{cybenko1989approximation,hornik1989multilayer} shows that fully-connected neural networks (FNNs) are capable to approximate the ground-truth classification function accurately, but in practice the performance of FNNs is far  from networks with specific architectures, such as the widely-used convolutional neural networks (CNN) \cite{krizhevsky2012imagenet} or the more recent capsule neural network (CapsNet) \cite{sabour2017dynamic}. The performance gap lies in the fact that the accuracy of NNs can be characterized by dividing the total error into three main types: approximation, optimization, and generalization \cite{bottou2008tradeoffs,lu2018collapse,jin2019quantifying,lu2019dying}. The universal approximation theorems only guarantee a small approximation error for a sufficiently large network, but they do not consider the optimization error and generalization error at all, which are equally important and often dominant contributions to the total error in practice. Useful networks should be easy to train, i.e., exhibit small optimization error, and generalize well to unseen data, i.e., exhibit small generalization error. 

To demonstrate the capability and effectiveness of learning nonlinear operators by neural networks, we setup the problem as general as possible by using the weakest possible constraints on the sensors and training dataset. Specifically, the only condition required is that the sensor locations $\{x_1, x_2, \dots, x_m\}$ are the same but not necessarily on a lattice for all input functions $u$, while we do not enforce any constraints on the output locations $y$ (Fig. \ref{fig:problem}B). To learn operators accurately and efficiently, we propose a specific network architecture, the deep operator network (DeepONet), to achieve smaller total error. We will demonsrate that the DeepONet significantly improves generalization based on a design of two sub-networks, the branch net for the input function and the trunk-net for the location to evaluate the output function.

We consider two types of operators, i.e., dynamic systems (e.g., in the form of ordinary differential equations, ODEs) and partial differential equations (PDEs). Dynamic systems are typically described by difference or differential equations, and identification of a nonlinear dynamic plant is a major concern in control theory. Some works~\cite{patra1999identification,zhao2009nonlinear} used neural networks to identify dynamic systems, but they only considered the dynamic systems described by difference equations. Some other works \cite{neofotistos2018machine,raissi2018multistep,qin2019data,erichson2019physics} predict the evolution of a specific dynamic system rather than identifying the system behavior for new unseen input signals. The network architectures they employed includes FNNs \cite{raissi2018multistep}, recurrent neural networks (RNNs) \cite{neofotistos2018machine}, reservoir computing \cite{neofotistos2018machine}, residual networks \cite{qin2019data}, autoencoder \cite{erichson2019physics}, neural ordinary differential equations \cite{chen2018neural}, and neural jump stochastic differential equations \cite{jia2019neural}. For identifying PDEs, some works treat the input and output function as an image, and then use CNNs to learn the image-to-image mapping \cite{winovich2019convpde,zhu2019physics}, but this approach can be only applied to the particular type of problems, where the sensors of the input function $u$ are distributed on a equispaced grid, and the training data must include all the $G(u)(y)$ values with $y$ also on a equispaced grid. In another approach without this restriction, PDEs are parametrized by unknown coefficients, and then only the coefficient values are identified from data \cite{brunton2016discovering,rudy2017data,zhang2019quantifying,pang2019fpinns,lu2019deepxde}. Alternatively, a generalized CNN based on generalized moving least squares \cite{trask2019gmls} can be used for unstructured data, but it can only approximate local operators and is not able to learn other operators like an integral operator.

The paper is organized as follows. In Section \ref{sec:method}, we present two network architectures of DeepONet: the stacked DeepONet and the unstacked DeepONet, and then introduce the data generation procedure. In Section \ref{sec:theory}, we present a theoretical analysis on the number of sensors required to represent the input function accurately for approximating ODE operators. In Section \ref{sec:result}, we test the performance of FNN, stacked DeepONet, and unstacked DeepONet for different examples, and demonstrate the accuracy and convergence rates of unstacked DeepONet. Finally, we conclude the paper in Section \ref{sec:conc}.

\section{Methodology}
\label{sec:method}
\subsection{Deep operator networks (DeepONets)}
\label{sec:deeponet}

We focus on learning operators in a more general setting, where the only requirement for the training dataset is the consistency of the sensors $\{x_1, x_2, \dots, x_m\}$ for input functions. In this general setting, the network inputs consist of two separate components:  $[u(x_1), u(x_2), \dots, u(x_m)]^T$ and $y$ (Fig. \ref{fig:problem}A), and the goal is to achieve good performance by designing the network architecture. One straightforward solution is to directly employ a classical network, such as FNN, CNN or RNN, and concatenate two inputs together as the network input, i.e., $[u(x_1), u(x_2), \dots, u(x_m), y]^T$. However, the input does not have any specific structure, and thus it is not meaningful to choose networks like CNN or RNN. Here we use FNN as the baseline model.

In high dimensional problems, $y$ is a vector with $d$ components, so the dimension of $y$ does not match the dimension of $u(x_i)$ for $i=1,2,\dots,m$ any more. This also prevents us from treating $u(x_i)$ and $y$ equally, and thus at least two sub-networks are needed to handle $[u(x_1), u(x_2), \dots, u(x_m)]^T$ and $y$ separately. Although the universal approximation theorem (Theorem \ref{thm:main}) does not have any guarantee on the total error, it still provides us a network structure in Eq. \ref{eq:thm}. Theorem \ref{thm:main} only considers a shallow network with one hidden layer, so we extend it to deep networks, which have more expressivity than shallow ones. The architecture we propose is shown in Fig. \ref{fig:problem}C, and the details are as follows. First there is a ``trunk'' network, which takes $y$ as the input and outputs $[t_1, t_2, \dots, t_p]^T \in \mathbb{R}^p$. In addition to the trunk network, there are $p$ ``branch'' networks, and each of them takes $[u(x_1), u(x_2), \dots, u(x_m)]^T$ as the input and outputs a scalar $b_k \in \mathbb{R}$ for $k=1,2,\dots,p$. We merge them together as in Eq. \ref{eq:thm}:
\[ G(u)(y) \approx \sum_{k=1}^p b_k t_k. \]
We note that the trunk network also applies activation functions in the last layer, i.e., $t_k = \sigma(\cdot)$ for $k=1,2,\dots,p$, and thus this trunk-branch network can also be seen as a trunk network with each weight in the last layer parameterized by another branch network instead of the classical single variable. We also note that in Eq. \ref{eq:thm} the last layer of each $b_k$ branch network does not have bias. Although bias is not necessary in Theorem \ref{thm:main}, adding bias may increase the performance by reducing the generalization error. In addition to adding bias to the branch networks, we may also add a bias $b_0 \in \mathbb{R}$ in the last stage:
\begin{equation}\label{eq:guy}
G(u)(y) \approx \sum_{k=1}^p b_k t_k + b_0.
\end{equation}

In practice, $p$ is at least of the order of 10, and using lots of branch networks is computationally and memory expensive. Hence, we merge all the branch networks into one single branch network (Fig. \ref{fig:problem}D), i.e., a single branch network outputs a vector $[b_1, b_2, \dots, b_p]^T \in \mathbb{R}^p$. In the first DeepONet (Fig. \ref{fig:problem}C), there are $p$ branch networks stacked parallel, so we name it as ``stacked DeepONet'', while we refer to the second DeepONet (Fig. \ref{fig:problem}D) as ``unstacked DeepONet''. All versions of DeepONets are implemented in DeepXDE \cite{lu2019deepxde}, a user-friendly Python library designed for scientific machine learning (\url{https://github.com/lululxvi/deepxde}).

DeepONet is a high-level network architecture without defining the architectures of its inner trunk and branch networks. To demonstrate the capability and good performance of DeepONet alone, we choose the simplest FNN as the architectures of the sub-networks in this study. It is possible that using convolutional layers we could further improve accuracy. However, convolutional layers usually work for square domains with $\{x_1, x_2, \dots, x_m\}$ on a equispaced grid, so as alternative and for a more general setting we may use the ``attention'' mechanism \cite{vaswani2017attention}.

Embodying some prior knowledge into neural network architectures usually induces good generalization. This inductive bias has been reflected in many networks, such as CNN for images and RNN for sequential data. The success of DeepONet even using FNN as its sub-networks is also due to its strong inductive bias. The output $G(u)(y)$ has two independent inputs $u$ and $y$, and thus using the trunk and branch networks explicitly is consistent with this prior knowledge. More broadly, $G(u)(y)$ can be viewed as a function of $y$ conditioning on $u$. Finding an effective way to represent the conditioning input is still an open question, and different approaches have been proposed, such as feature-wise transformations \cite{dumoulin2018feature-wise}.

\subsection{Data generation}

The input signal $u(x)$ of the process plays an important role in system identification. Clearly, the input signal is the only possibility to influence the process in order to gather information about its response, and the quality of the identification signal determines an upper bound on the accuracy that in the best case can be achieved by any model. In this study, we mainly consider two function spaces: Gaussian random field (GRF) and orthogonal (Chebyshev) polynomials.

We use the mean-zero GRF:
$$u \sim \mathcal{G}(0, k_l(x_1, x_2)),$$
where the covariance kernel $k_l(x_1, x_2) = \exp(-\| x_1 -x_2 \|^2 / 2l^2)$ is the radial-basis function (RBF) kernel with a length-scale parameter $l > 0$. The length-scale $l$ determines the smoothness of the sampled function, and larger $l$ leads to smoother $u$.

Let $M>0$ and $T_i$ are Chebyshev polynomials of the first kind. We define the orthogonal polynomials of degree $N$ as: $$V_{\text{poly}}=\left\{ \sum_{i=0}^{N-1} a_i T_i(x): |a_i| \leq M \right\}.$$
We generate the dataset from $V_{\text{poly}}$ by randomly sampling $a_i$ from $[-M,M]$ to get a sample of $u$.

After sampling $u$ from the chosen function spaces, we solve the ODE systems by  Runge-Kutta (4, 5) and PDEs by a second-order finite difference method to obtain the reference solutions. We note that one data point is a triplet $(u, y, G(u)(y))$, and thus one specific input $u$ may appear in multiple data points with different values of $y$. For example, a dataset of size 10000 may only be generated from 100 $u$ trajectories, and each evaluates $G(u)(y)$ for 100 $y$ locations.

\section{Number of sensors for identifying nonlinear dynamic systems}
\label{sec:theory}

In this section, we investigate how many sensor points we need to achieve accuracy $\varepsilon$ for identifying nonlinear dynamic systems. Suppose that the dynamic system is subject to the following ODE system:
\begin{equation} \label{eq:ODE_system}
\left\{\begin{array}{l}\frac{d}{dx}\boldsymbol{s}(x)=\boldsymbol{g}(\boldsymbol{s}(x),u(x),x) \\
\boldsymbol{s}(a)=\boldsymbol{s_0}\end{array}\right. ,
\end{equation}
where $u\in V$(a compact subset of $C[a,b]$) is the input signal, and $\boldsymbol{s}:[a,b]\rightarrow\mathbb{R}^{K}$ is the solution of system (\ref{eq:ODE_system}) serving as the output signal.

Let $G$ be the operator mapping the input $u$ to the output $\boldsymbol{s}$, i.e., $Gu$ satisfies
\begin{equation*}\label{eq:Gu}
    (Gu)(x)=\boldsymbol{s_0}+\int_a^x\boldsymbol{g}((Gu)(t),u(t),t)dt.
\end{equation*}
Now, we choose uniformly $m+1$ points $x_{j}=a+j(b-a)/m,j=0,1,\cdots,m$ from $[a,b]$, and define the function $u_{m}(x)$ as follows:
\begin{equation*}
    u_m(x)=u(x_j)+\frac{u(x_{j+1})-u(x_j)}{x_{j+1}-x_j} (x-x_j),\ x_j\leq x\leq x_{j+1},\ j=0,1,\cdots,m-1.
\end{equation*}
Denote the operator mapping $u$ to $u_{m}$ by $\mathcal{L}_{m}$, and let $U_{m}=\{\mathcal{L}_{m}(u)|u\in V\}$, which is obviously a compact subset of $C[a,b]$ since $V$ is compact and continuous operator $\mathcal{L}_{m}$  keeps the compactness. Naturally, $W_{m}\coloneqq V\cup U_{m}$ as the union of two compact sets is also compact. Then, set $W\coloneqq\bigcup_{i=1}^{\infty}W_{i}$, and Lemma \ref{lem:W_compact} points out that $W$ is still a compact set.
Since $G$ is a continuous operator,  $G(W)$ is compact in $C([a,b];\mathbb{R}^{K})$.
The subsequent discussions are mainly within $W$ and $G(W)$. For convenience of analysis, we assume that $\boldsymbol{g}(\boldsymbol{s},u,x)$ satisfies the Lipschitz condition with respect to $\boldsymbol{s}$ and $u$ on $G(W)\times W$, i.e., there is a constant $c>0$ such that
\begin{equation*}
    \begin{array}{l}
    \|\boldsymbol{g}(\boldsymbol{s_1},u,x)-\boldsymbol{g}(\boldsymbol{s_2},u,x)\|_{2}\leq c\|\boldsymbol{s_1}-\boldsymbol{s_2}\|_{2}\\
    \|\boldsymbol{g}(\boldsymbol{s},u_1,x)-\boldsymbol{g}(\boldsymbol{s},u_2,x)\|_{2}\leq c|u_1-u_2|
    \end{array}.
\end{equation*}
Note that this condition is easy to achieve, for instance, as long as $\boldsymbol{g}$ is differentiable with respect to $\boldsymbol{s}$ and $u$ on $G(W)\times W$.

For $u\in V,u_{m}\in U_{m}$, there exists a constant $\kappa(m,V)$ depending on $m$ and compact space $V$, such that
\begin{equation} \label{eq:max_bound}
    \max_{x\in[a,b]}|u(x)-u_m(x)|\leq\kappa(m,V),\quad \kappa(m,V)\rightarrow 0\ \ as\ \ m\rightarrow\infty.
\end{equation}
When $V$ is GRF with RBF kernel, we have $\kappa(m,V) \sim \frac{1}{m^2l^2}$, see Appendix \ref{apd:grf} for the proof.
Based on the these concepts, we have the following theorem.

\begin{theorem}\label{thm:m}
Suppose that $m$ is a positive integer making $c(b-a)\kappa(m,V)e^{c(b-a)}$ less than $\varepsilon$, then for any $d\in [a,b]$, there exist $\mathcal{W}_{1}\in \mathbb{R}^{n\times (m+1)},b_{1}\in \mathbb{R}^{m+1},\mathcal{W}_{2}\in \mathbb{R}^{K\times n},b_{2}\in \mathbb{R}^{K}$, such that
\begin{equation*}
    \|(Gu)(d)-(\mathcal{W}_{2}\cdot\sigma(\mathcal{W}_{1}\cdot [u(x_{0})\quad\cdots\quad u(x_{m})]^{T}+b_{1})+b_{2})\|_{2}<\varepsilon
\end{equation*} 
holds for all $u\in V$.
\end{theorem}
\begin{proof}
The proof can be found in Appendix \ref{apd:m}.
\end{proof}

\section{Simulation results}
\label{sec:result}

In this section, we first show that DeepONets have better performance than FNNs due to the smaller generalization error even in the easiest linear problem, and then demonstrate the capability of DeepONets for three nonlinear ODE and PDE problems. In all problems, we use the Adam optimizer with learning rate 0.001, and the number of iterations is chosen to guarantee the training is convergent. The other parameters and network sizes are listed in Tables \ref{tab:par} and \ref{tab:DeepONet}, unless otherwise stated. The codes of all examples are published in GitHub (\url{https://github.com/lululxvi/deepxde}).

\begin{table}[htbp]
\centering
\caption{\textbf{Default parameters for each problem, unless otherwise stated.} We note that one data point is a triplet $(u, y, G(u)(y))$, and thus one specific input $u$ may generate multiple data points with different values of $y$.}
\label{tab:par}
\begin{tabular}{c|cccccc}
\toprule
Case & $u$ space & \# Sensors $m$ & \# Training & \# Test & \# Iterations & Other parameters \\
\midrule
4.1.1 & GRF ($l=0.2$) & 100 & 10000 & 100000 & 50000 & \\
4.1.2 & GRF ($l=0.2$) & 100 & 10000 & 100000 & 100000 & \\
4.2 & GRF ($l=0.2$) & 100 & 10000 & 100000 & 100000 & $k=1, T=1$ \\
4.3 & GRF ($l=0.2$) & 100 &  & 1000000 & 500000 & \\
\bottomrule
\end{tabular}
\end{table}

\begin{table}[htbp]
\centering
\caption{\textbf{DeepONet size for each problem, unless otherwise stated.}}
\label{tab:DeepONet}
\begin{tabular}{c|cccccc}
\toprule
Case & Network type & Trunk depth & Trunk width & Branch depth & Branch width  \\
\midrule
4.1 & Stacked/Unstacked & 3 & 40 & 2 & 40 \\
4.2 & Unstacked & 3 & 40 & 2 & 40 \\
4.3 & Unstacked & 3 & 100 & 2 & 100 \\
\bottomrule
\end{tabular}
\end{table}

\subsection{A simple 1D dynamic system}

A 1D dynamic system is described by
$$\frac{ds(x)}{dx} = g(s(x), u(x), x), \quad x \in [0, 1],$$
with an initial condition $s(0)=0$. Our goal is to predict $s(x)$ over the whole domain $[0, 1]$ for any $u(x)$.

\subsubsection{Linear case: $g(s(x), u(x), x) = u(x)$}
\label{sec:antiderivative}

\begin{figure}[htbp]
\centering
\includegraphics{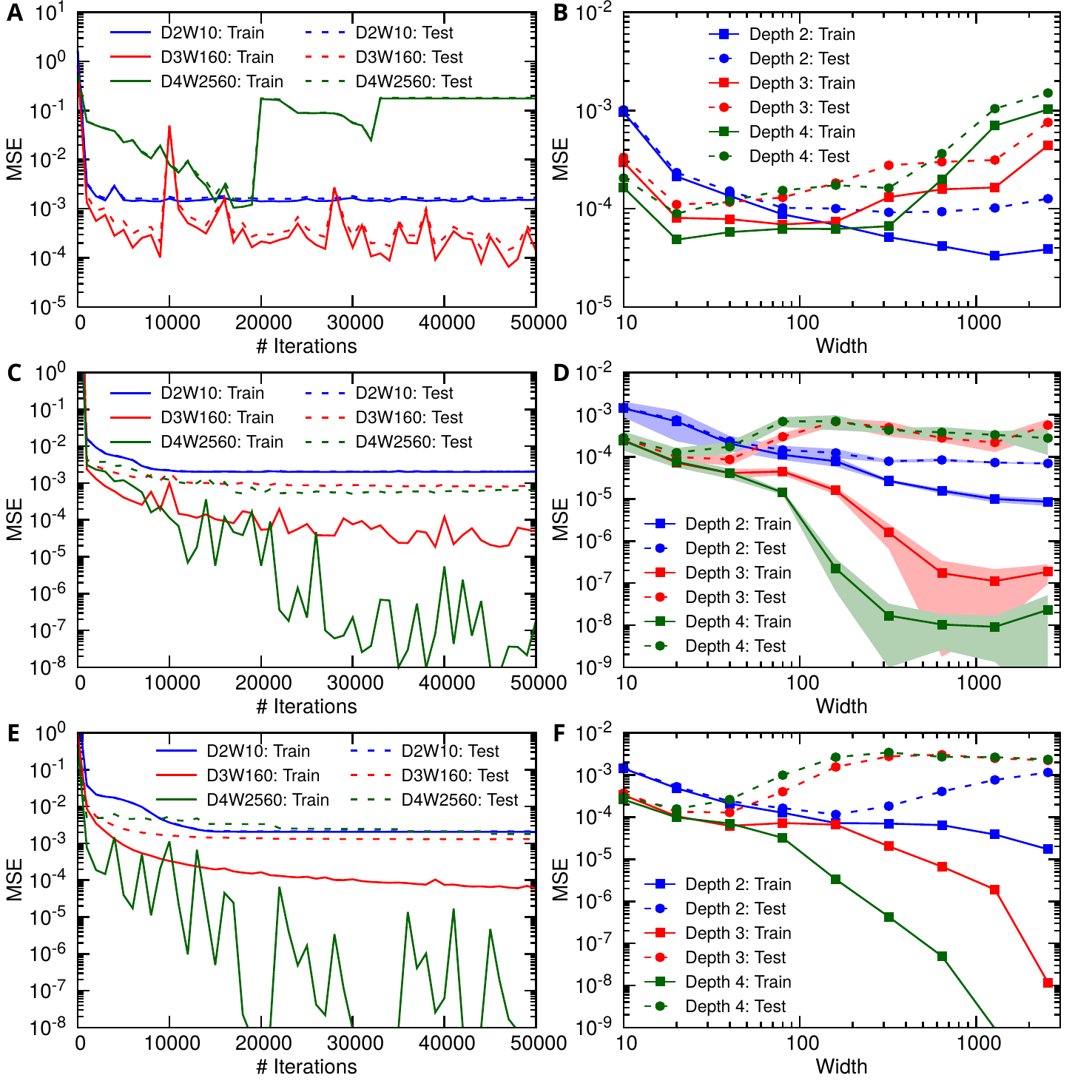}
\caption{\textbf{Errors of FNNs trained to learn the antiderivative operator (linear case).} (\textbf{A} and \textbf{B}) Learning rate 0.01. (\textbf{C} and \textbf{D}) Learning rate 0.001. (\textbf{E} and \textbf{F}) Learning rate 0.0001. (A, C, E) The solid and dash lines are the training error and test error during the training process, respectively. The blue, red and green lines represent FNNs of size (depth 2, width 10), (depth 3, width 160), and (depth 4, width 2560), respectively. (B, D, F) The blue, red and green lines represent FNNs of depth 2, 3 and 4, respectively. The shaded regions in (D) are the one-standard-derivation (SD) from 10 runs with different training/test data and network initialization. For clarity, only the SD of learning rate 0.001 is shown (Fig. D). The number of sensors for $u$ is $m=100$.}
\label{fig:fnn}
\end{figure}

We first consider a linear problem by choosing $g(s(x), u(x), x) = u(x)$, which is equivalent to learning the antiderivative operator
$$G: u(x) \mapsto s(x) = \int_0^x u(\tau) d\tau.$$
As the baseline, we train FNNs to learn the antiderivative operator. To obtain the best performance of FNNs, we grid search the three hyperparameters: depth from 2 to 4, width from 10 to 2560, and learning rate from 0.0001 to 0.01. The mean squared error (MSE) of the test dataset with learning rate 0.01, 0.001, and 0.0001 are shown in Fig. \ref{fig:fnn}. Although we only choose depth up to 4, the results show that increasing the depth further does not improve the error. Among all these hyperparameters, the smallest test error $\sim 10^{-4}$ is obtained for the network with depth 2, width 2560, and learning rate 0.001. We observe that when the network is small, the training error is large and the generalization error (the difference between test error and training error) is small, due to small expressivity. When the network size increases, the training error decreases, but the generalization error increases. We note that FNN has not reached the overfitting region, where the test error increases.

\begin{figure}[htbp]
\centering
\includegraphics{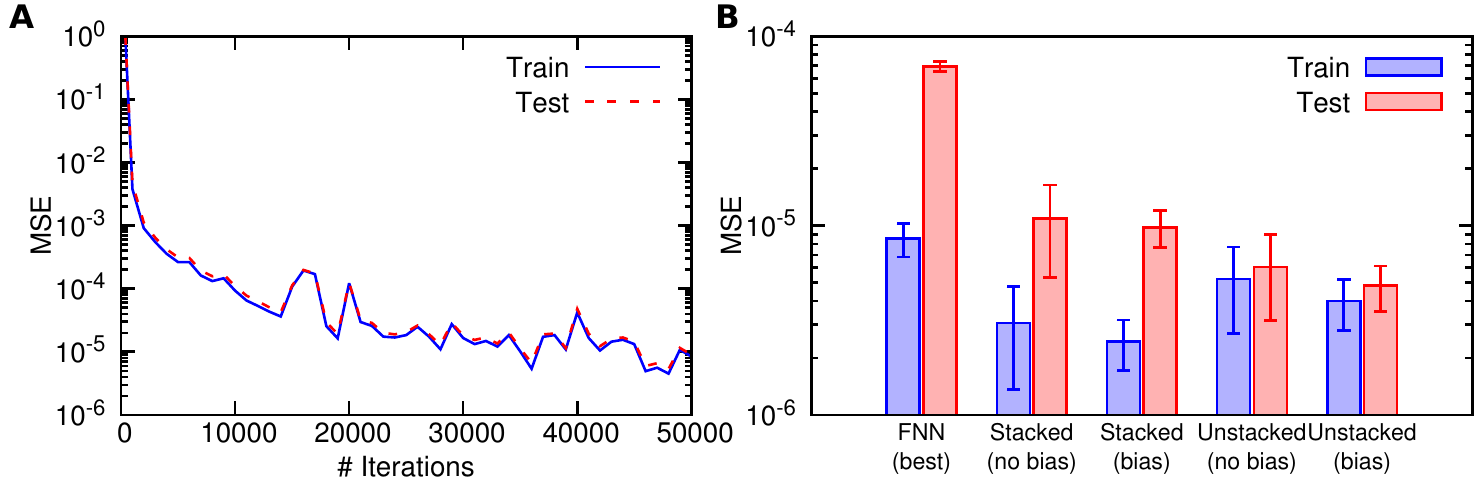}
\caption{\textbf{Errors of DeepONets trained to learn the antiderivative operator (linear case).} (\textbf{A}) The training trajectory of an unstacked DeepONet with bias. (\textbf{B}) The training/test error for stacked/unstacked DeepONets with/without bias compared to the best error of FNNs. The error bars are the one-standard-derivation from 10 runs with different training/test data and network initialization.}
\label{fig:coefnet}
\end{figure}

Compared to FNNs, DeepONets have much smaller generalization error and thus smaller test error. Here we do not aim to find the best hyperparameters, and only test the performance of the two DeepONets listed in Table \ref{tab:DeepONet}. The training trajectory of an unstacked DeepONet with bias is shown in Fig. \ref{fig:coefnet}A, and the generalization error is negligible. We observe that for both stacked and unstacked DeepONets, adding bias to branch networks and Eq.~(\ref{eq:guy}) reduces both training and test errors (Fig. \ref{fig:coefnet}B); DeepONets with bias also have smaller uncertainty, i.e., more stable for training from random initialization (Fig. \ref{fig:coefnet}B). Compared to stacked DeepONets, although unstacked DeepONets have larger training error, the test error is smaller, due to the smaller generalization error. Therefore, unstacked DeepONets with bias achieve the best performance. In addition, unstacked DeepONets have fewer number of parameters than stacked DeepONets, and thus can be trained faster using much less memory.

\subsubsection{Nonlinear case: $g(s(x), u(x), x) = -s^2(x) + u(x)$}

\begin{figure}[htbp]
\centering
\includegraphics{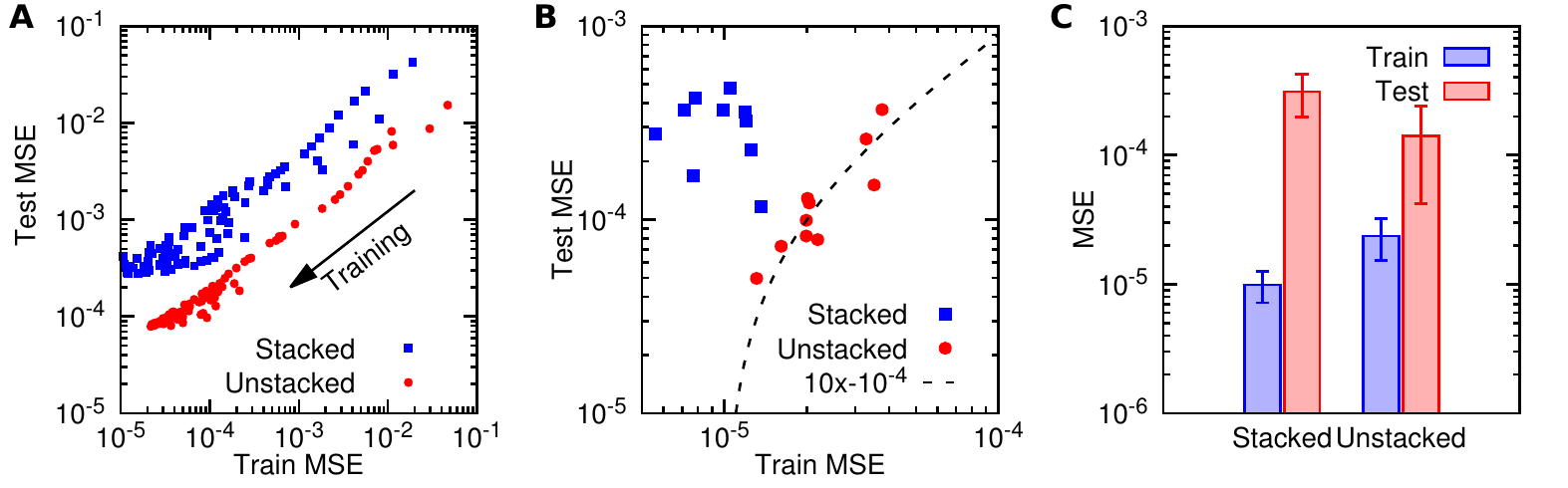}
\caption{\textbf{Nonlinear ODE: unstacked DeepONets have smaller generalization error and smaller test MSE than stacked DeepONets.} (\textbf{A}) The correlation between the training MSE and the test MSE of one stacked/unstacked DeepONet during the training process. (\textbf{B}) The correlation between the final training MSE and test MSE of stacked/unstacked DeepONets in 10 runs with random training dataset and network initialization. The training and test MSE of unstacked DeepONets follow a linear correlation (black dash line). (\textbf{C}) The mean and one-standard-derivation of data points in B.}
\label{fig:ode_stacked}
\end{figure}

\begin{figure}[htbp]
\centering
\includegraphics{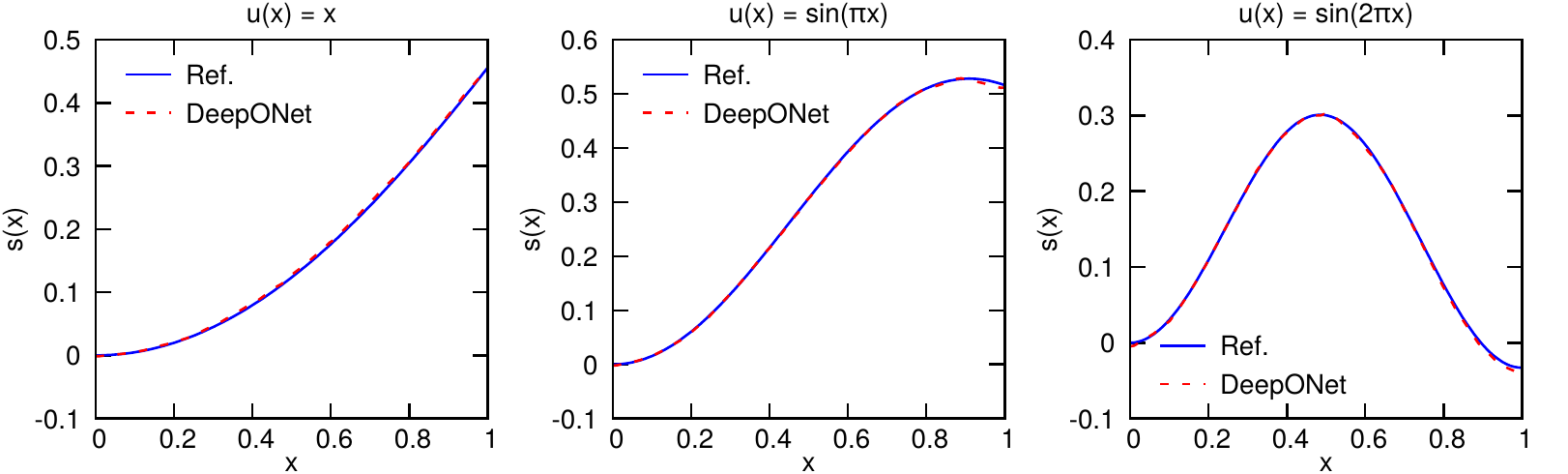}
\caption{\textbf{Nonlinear ODE: predictions of a trained unstacked DeepONet on three out-of-distribution input signals.} The blue and red lines represent the reference solution and the prediction of a DeepONet.}
\label{fig:ode_example}
\end{figure}

Next we consider a nonlinear problem with $g(s(x), u(x), x) = -s^2(x) + u(x)$. Because $s(x)$ may explode for certain $u$, we compute the test MSE by removing the 1\textperthousand\ worst predictions. During the network training, the training MSE and test MSE of both stacked and unstacked DeepONets decrease, but the correlation between training MSE and test MSE of unstacked DeepONets is tighter (Fig. \ref{fig:ode_stacked}A), i.e., smaller generalization error. This tight correlation between training and test MSE of unstacked DeepONets is also observed across multiple runs with random training dataset and network initialization (Fig. \ref{fig:ode_stacked}B). Moreover, the test MSE and training MSE of unstacked DeepONets follow almost a linear correlation
$$\text{MSE}_{\text{test}} \approx 10 \times \text{MSE}_{\text{train}} - 10^{-4}.$$
Fig. \ref{fig:ode_stacked}C shows that unstacked DeepONets have smaller test MSE due to smaller generalization error. DeepONets work even for out-of-distribution predictions, see three examples of the prediction in Fig. \ref{fig:ode_example}. In the following study, we will use unstacked DeepONets.

\subsection{Gravity pendulum with an external force}

The motion of a gravity pendulum with an external force is described as
\begin{align*}
\frac{ds_1}{dt} &= s_2, \\
\frac{ds_2}{dt} &= -k\sin s_1 + u(t),
\end{align*}
with an initial condition $\mathbf{s}(0) = \mathbf{0}$, and $k$ is determined by the acceleration due to gravity and the length of the pendulum. This problem is characterized by three factors: (1) $k$, (2) maximum prediction time $T$, and (3) input function space. The accuracy of learned networks is determined by four factors: (1) the number of sensor points $m$; (2) training dataset size; (3) network architecture, (4) optimizer. We investigate the effects of these factors on the accuracy.

\subsubsection{Number of sensors}
\label{sec:sensor}

\begin{figure}[htbp]
\centering
\includegraphics{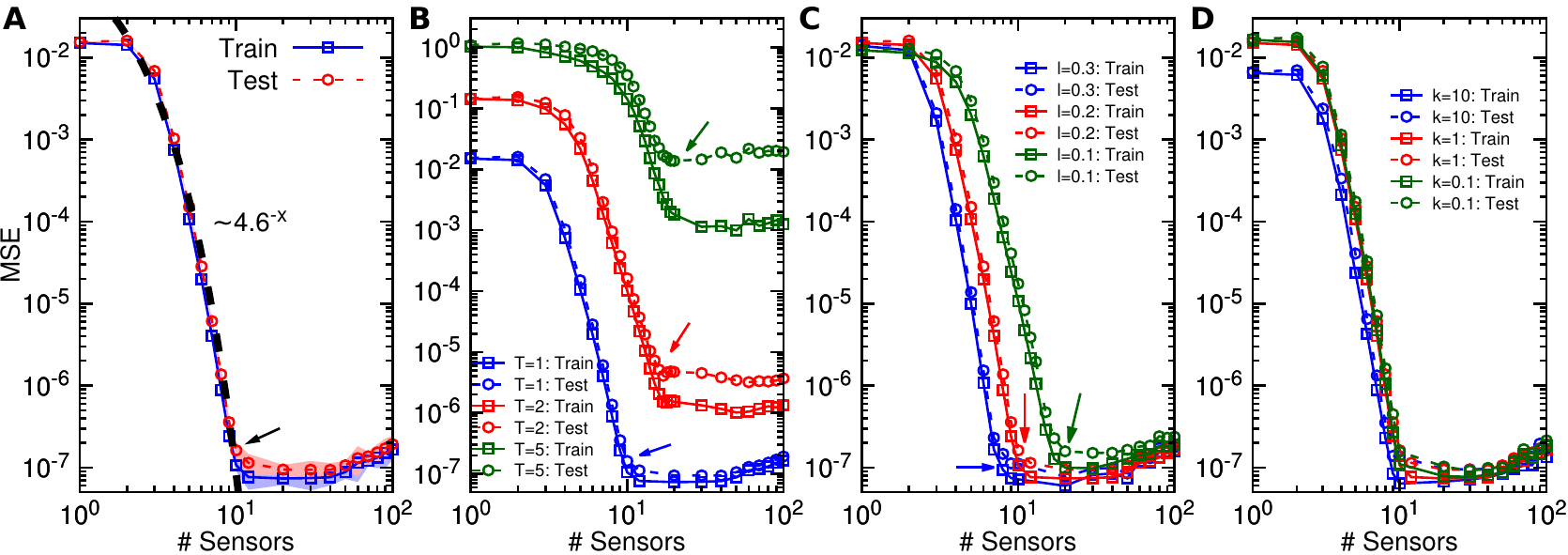}
\caption{\textbf{Gravity pendulum: required number of sensors for different $T$, $k$ and $l$.} (\textbf{A}) Training MSE (square symbols) and test MSE (circle symbols) decrease as the number of sensors increases in the case $k=1$, $T=1$ and $l=0.2$. Training and test MSE versus the number of sensors in different conditions of (\textbf{B}) $T$, (\textbf{C}) $l$, and (\textbf{D}) $k$. For clarity, the SD is not shown. The arrow indicates where the rate of the error decay diminishes.}
\label{fig:m}
\end{figure}

The number of sensors required to distinguish two input functions depends on the value of $k$, prediction time $T$, and the input function space. For the case with $k=1$, $T=1$ and $l=0.2$, when the number of sensors $m$ is small, the error decays exponentially as we increase the  number  of sensors, Fig. \ref{fig:m}A):
\[ \text{MSE} \propto 4.6^{-\# \text{sensors}}. \]
When $m$ is already large, the effect of increasing $m$ is negligible. The transition occurs at $\sim$10 sensors, as indicated by the arrow.

To predict $s$ for a longer time, more sensors are required (Fig. \ref{fig:m}B). For example, predicting until $T=5$ requires $\sim$25 sensors. If the function $u$ is less smooth corresponding to smaller $l$, it also requires more sensors (Fig. \ref{fig:m}C). However, the number of sensors is not sensitive to $k$ (Fig. \ref{fig:m}D). Although it is hard to quantify the exact dependency of $m$ on $T$ and $l$, by fitting the computational results we show that
$$m \propto \sqrt{T} \quad \text{and} \quad m \propto l^{-1}.$$
In Theorem \ref{thm:m}, $m$ should be large enough to make $Te^{cT}\kappa(m,V)$ small. In Appendix \ref{apd:grf} we show theoretically that $m \propto l^{-1}$ for the GRF function space with RBF kernel, which is consistent with our computational result here. $Te^{cT}$ in the bound is loose compared to the computational results.

\subsubsection{Error tendency and convergence}
\label{sec:error}

\begin{figure}[htbp]
\centering
\includegraphics{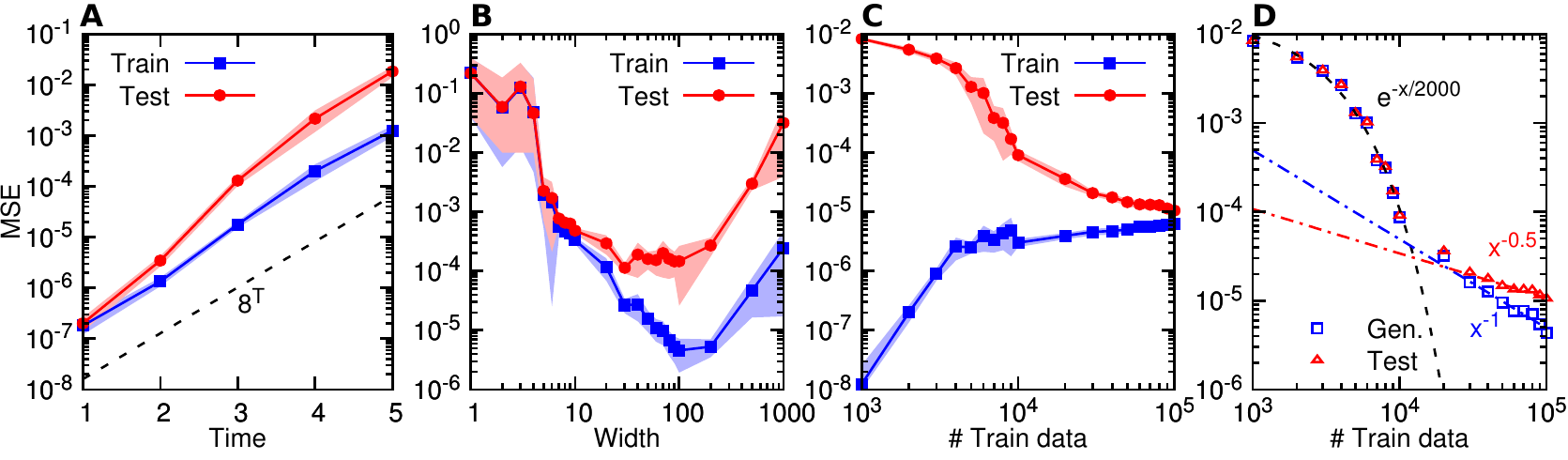}
\caption{\textbf{Gravity pendulum: error tendency and convergence.} (\textbf{A}) Both training error (blue) and test error (red) increase fast for long-time prediction if we keep the network size fixed at width 40 and the training data size fixed at 10000 points. (\textbf{B}) Training and test errors first decrease and then increase, when network width increases ($T=3$). (\textbf{C}) Test error decreases when more training data are used ($T=3$, width 100). (\textbf{D}) Test error and generalization error have exponential convergence for small training dataset, and then converge with rate $x^{-0.5}$ and $x^{-1}$, respectively ($T=3$, width 100).}
\label{fig:t}
\end{figure}

Here we investigate the error tendency under different conditions, including prediction time, network size, and training dataset size. We first observe that both the training and test errors grow exponentially with the maximum prediction time (Fig. \ref{fig:t}A): 
\[ \text{MSE} \propto 8^T. \]
We note that the error is not the error at time $T$ but the average error over the domain $[0, T]$, because we predict the $s(t)$ for any $t \in [0, T]$. As shown in Fig. \ref{fig:m}B, 100 sensor points are sufficient for $T=5$, and thus the possible reasons for increased error are: (1) the network is too small, and (2) training data is not sufficient. Because the error in $T=1$ is already very small, to leverage the error, we use $T=3$ in the following experiments. By varying the network width, we can observe that there is a best width to achieve the smallest error (Fig. \ref{fig:t}B). It is reasonable that increasing the width from 1 to 100 would decrease the error, but the error would instead increase when the width further increases. This could be due to the increased optimization error, and a better learning rate may be used to find a better network.

To examine the effect of training dataset size, we choose networks of width 100 to eliminate the network size effect. The training, test, and generalization errors using different dataset size are shown in Fig. \ref{fig:t}C and D, and we have
\begin{equation*}
\text{MSE}_{\text{test}} \propto \begin{cases} e^{-x/2000}, & \text{for small dataset} \\ x^{-0.5}, & \text{for large dataset} \end{cases}, \quad
\text{MSE}_{\text{gen}} \propto \begin{cases} e^{-x/2000}, & \text{for small dataset} \\ x^{-1}, & \text{for large dataset} \end{cases},
\end{equation*}
where $x$ is the number of training data points. It is surprising that test error and generalization error have exponential convergence for training dataset size $< 10^4$. Even for large dataset, the convergence rate of $x^{-1}$ for the generalization error is still higher than the classical $x^{-0.5}$ in the learning theory \cite{mitzenmacher2017probability}. This fast convergence confirms the exceptional performance of DeepONets, especially in the region of small dataset.

\subsubsection{Input function spaces}

\begin{figure}[htbp]
\centering
\includegraphics{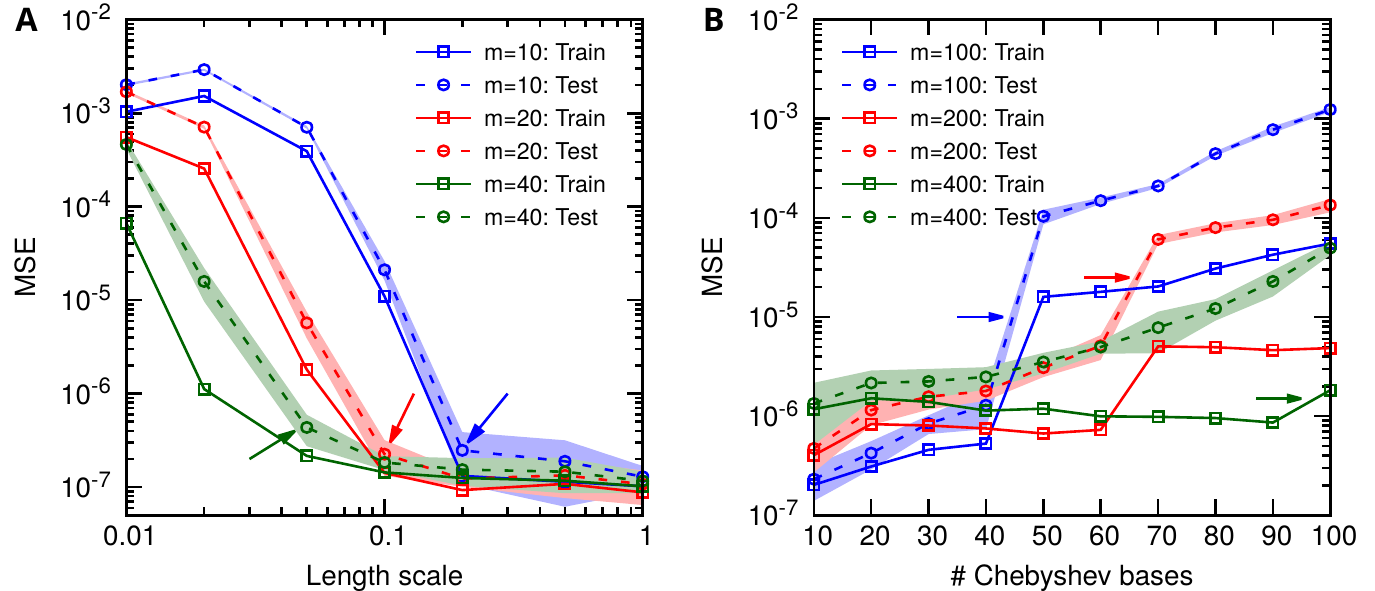}
\caption{\textbf{Gravity pendulum: errors for different input function spaces.} Training error (solid line) and test error (dash line) for (\textbf{A}) GRF function space with different length scale $l$. The colors correspond to the number of sensors as shown in the inset. (\textbf{B}) Chebyshev polynomials for different number of basis functions.}
\label{fig:input}
\end{figure}

Next we investigate the effect of different function spaces, including GRF with different length scale $l$ and the space of Chebyshev polynomials with different number of bases. For a fixed sensor number, we observe that there exists a critical length scale, around where the training and test errors change rapidly, see the arrows in Fig. \ref{fig:input}A and B. This sharp transition around the critical value is also observed in the space of Chebyshev polynomials with different number of bases (Fig. \ref{fig:input}B). The relations between the critical values and the sensor numbers are
\[ l \propto m^{-1} \quad \text{and} \quad \# \text{Bases} \propto \sqrt{m}. \]
The inverse relation between $l$ and $m$ is consistent with the theoretical results in Appendix \ref{apd:grf} and the computational results in Section \ref{sec:sensor}. Therefore, when the function space complexity increases, one may increase $m$ to capture the functions well.

\subsection{Diffusion-reaction system with a source term}

\begin{figure}[htbp]
\centering
\includegraphics{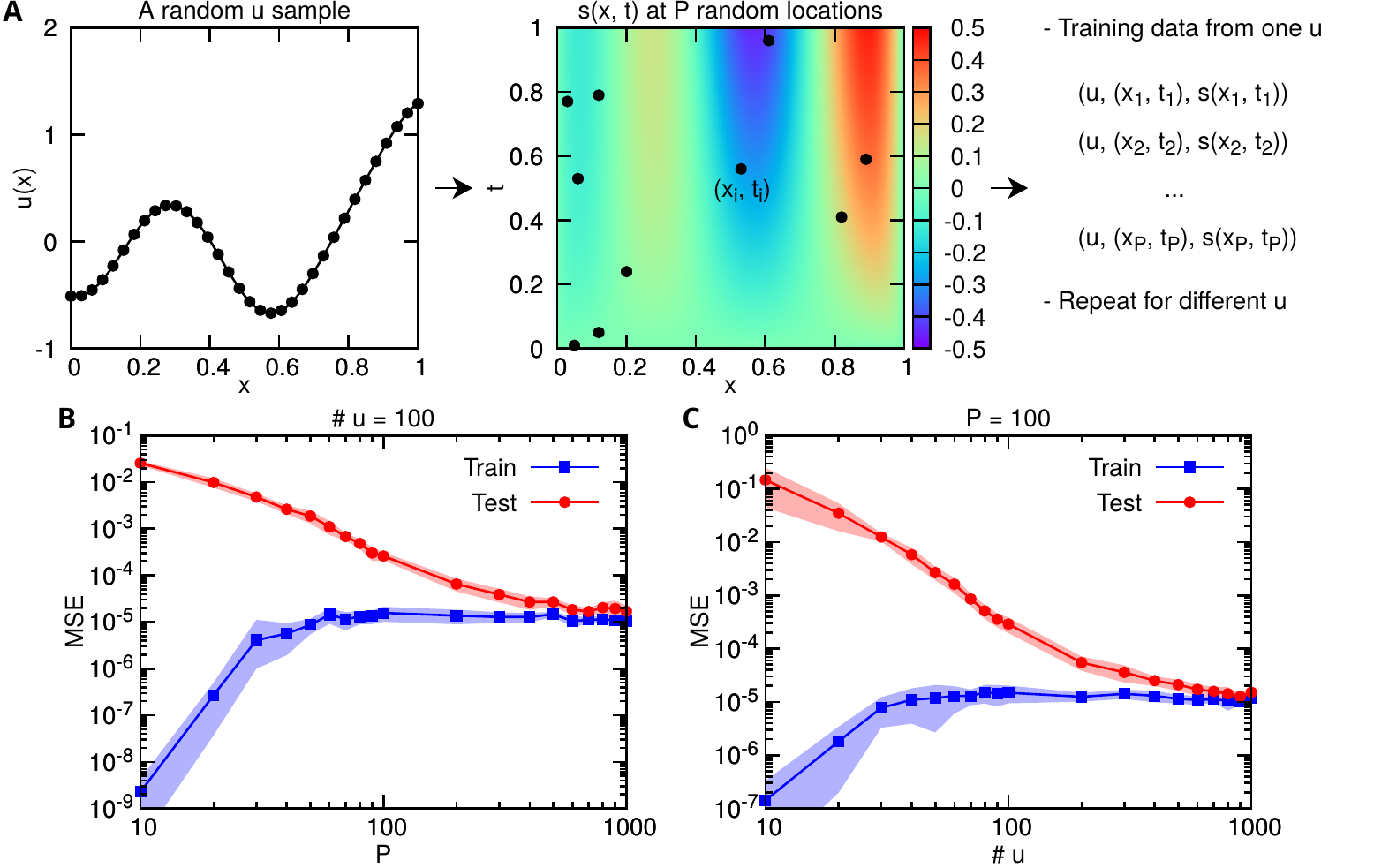}
\caption{\textbf{Learning a diffusion-reaction system.} (\textbf{A}) (left) An example of a random sample of the input function $u(x)$. (middle) The corresponding output function $s(x,t)$ at $P$ different $(x,t)$ locations. (right) Pairing of inputs and outputs at the training data points. The total number of training data points is the product of $P$ times the number of samples of $u$. (\textbf{B}) Training error (blue) and test error (red) for different values of the number of random points $P$ when 100 random $u$ samples are used. (\textbf{C}) Training error (blue) and test error (red) for different number of $u$ samples when $P=100$. The shaded regions denote one-standard-derivation.}
\label{fig:adr_error}
\end{figure}

A diffusion-reaction system with a source term $u(x)$ is described by
\begin{equation*}
    \frac{\partial s}{\partial t} = D\frac{\partial^2 s}{\partial x^2} + k s^2 + u(x), \quad x \in [0, 1], t \in [0, 1],
\end{equation*}
with zero initial/boundary conditions, where $D=0.01$ is the diffusion coefficient, and $k=0.01$ is the reaction rate. We use DeepONets to learn the operator mapping from $u(x)$ to the PDE solution $s(x, t)$. In the previous examples, for each input $u$, we only use one random point of $s(x)$ for training, and instead we may also use multiple points of $s(x)$. To generate the training dataset, we solve the diffusion-reaction system using a second-order implicit finite difference method on a 100 by 100 grid, and then for each $s$ we randomly select $P$ points out of these $10000=100 \times 100$ grid points (Fig. \ref{fig:adr_error}A). Hence, the dataset size is equal to the product of $P$ by the number of $u$ samples. We confirm that the training and test datasets do not include the data from  the same $s$.

We investigate the error tendency with respect to (w.r.t.) the number of $u$ samples and the value of $P$. When we use 100 random $u$ samples, the test error decreases first as $P$ increases (Fig. \ref{fig:adr_error}B), and then saturates due to other factors, such as the finite number of $u$ samples and fixed neural network size. We observe a similar error tendency but with less saturation as the number of $u$ samples increases with $P$ fixed (Fig. \ref{fig:adr_error}C). In addition, in this PDE problem the DeepONet is able to learn from a small dataset, e.g., a DeepONet can reach the test error of $\sim 10^{-5}$ when it is only trained with 100 $u$ samples ($P=1000$). We recall that we test on 10000 grid points, and thus on average each location point only has $100 \times 1000 / 10000 = 10$ training data points.

\begin{figure}[htbp]
\centering
\includegraphics{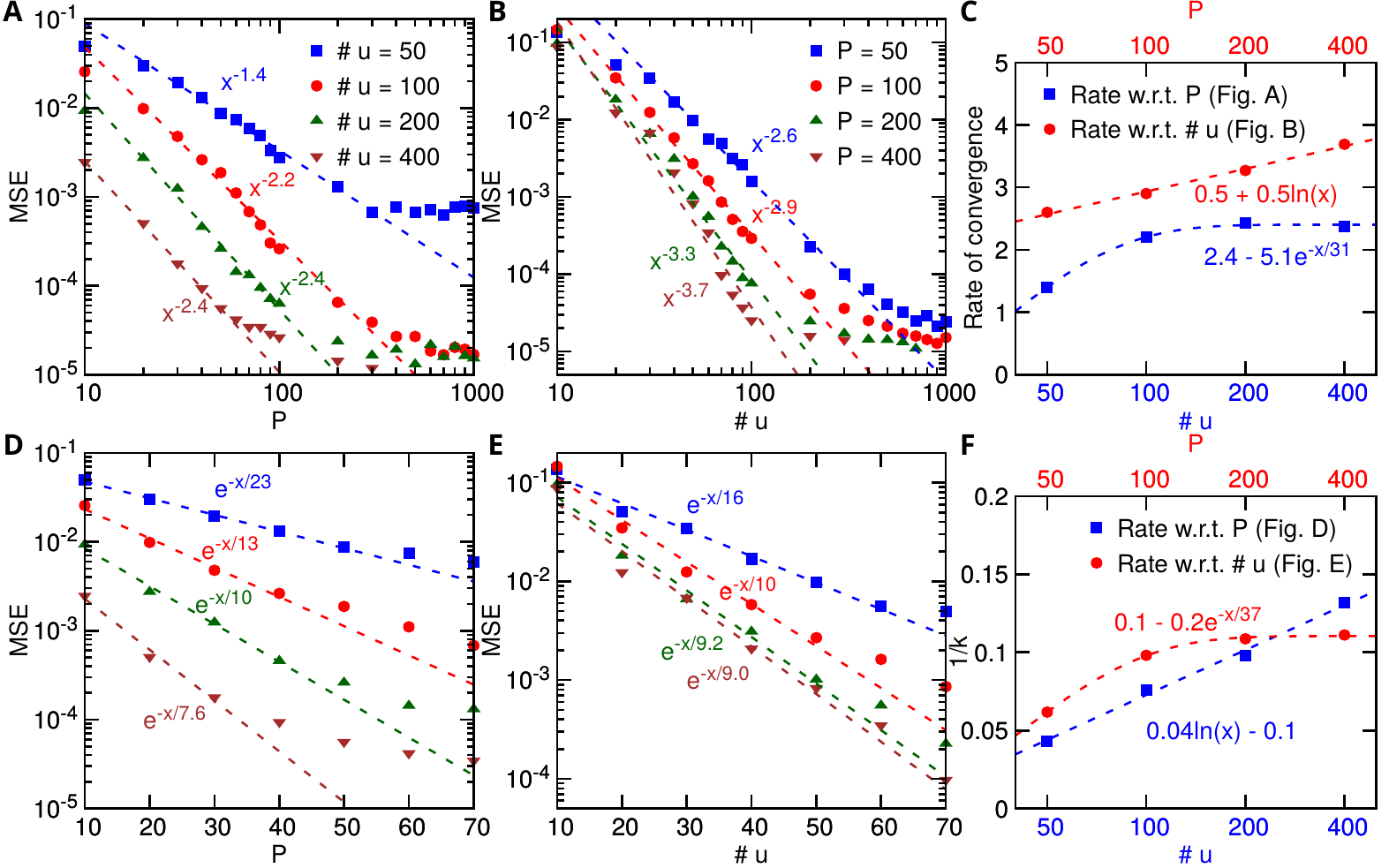}
\caption{\textbf{Error convergence rates for different number of training data points.} (\textbf{A}) Convergence of test error with respect to $P$ for different number of $u$ samples. (\textbf{B}) Convergence of test error with respect to the number of $u$ samples for different values of $P$. (\textbf{C}) The polynomial rates of convergence versus the number of $u$ samples or the values of $P$. (\textbf{D}) Exponential convergence of test error with respect to $P$ for different number of $u$ samples. (\textbf{E}) Exponential convergence of test error with respect to the number of $u$ samples for different values of $P$. (\textbf{F}) The coefficient $1/k$ in the exponential convergence $e^{-x/k}$ versus the number of $u$ samples or the values of $P$.}
\label{fig:adr_convergence}
\end{figure}

Before the error saturates, the rates of convergence w.r.t. both $P$ and the number of $u$ samples obey a polynomial law in the most of the range (Figs. \ref{fig:adr_convergence}A and B). The rate of convergence w.r.t. $P$ depends on the number of $u$ samples, and more $u$ samples induces faster convergence until it saturates (the blue line in Fig. \ref{fig:adr_convergence}C). Similarly, the rate of convergence w.r.t. the number of $u$ samples depends on the value of $P$ (the red line in Fig. \ref{fig:adr_convergence}C). In addition, in the initial range of the convergence, we observe an exponential convergence (Figs. \ref{fig:adr_convergence}D and E) as in Section \ref{sec:error}. The coefficient $1/k$ in the exponential convergence $e^{-x/k}$ also depends on the number of $u$ samples or the value of $P$ (Fig. \ref{fig:adr_convergence}F). It is reasonable that the convergence rate in Figs. \ref{fig:adr_convergence}C and F increases with the number of $u$ samples or the value of $P$, because the total number of training data points is equal to $P\times \#u$. However, by fitting the points, it is surprising that there is a clear tendency in the form of either $\ln(x)$ or $e^{-x}$, which we cannot fully explain yet, and hence more theoretical and computational investigations are required.

\section{Conclusion}
\label{sec:conc}

In this paper, we formulate the problem of learning operators in a more general setup, and propose DeepONets to learn nonlinear operators. In DeepONets, we first construct two sub-networks to encode input functions and location variables separately, and then merge them together to compute the output. We test DeepONets on four ordinary/partial differential equation problems, and show that DeepONets can achieve small generalization errors by employing this inductive bias. In our simulations, we study systematically the effects on the test error of different factors, including the number of sensors, maximum prediction time, the complexity of the space of input functions, training dataset size, and network size. We observe different order polynomial and even exponential error convergence with respect to the training dataset size. To the best of our knowledge, this is the first time exponential convergence is observed in deep learning. Moreover, we derive theoretically the dependence of approximation error on different factors, which is consistent with our computational results.

Despite the aforementioned achievements, more work should be done both theoretically and computationally. For example, there have not been any theoretical results of network size for operator approximation, similar to the bounds of width and depth for function approximation \cite{hanin2017universal}. We also do not understand theoretically yet why DeepONets can induce small generalization errors. On the other hand, in this paper we use fully-connected neural networks for the two sub-networks, but as we discussed in Section \ref{sec:deeponet}, we can also employ other network architectures, such as convolutional neural networks or ``attention'' mechanism. These modifications may improve further the accuracy of DeepONets.

\section{Acknowledgments}

We thank Yanhui Su of Fuzhou University for the help on Theorem \ref{thm:m}. We thank Zhongqiang Zhang of Worcester Polytechnic Institute for the proof in Appendix \ref{apd:grf}. This work is supported by the DOE PhILMs project (No. de-sc0019453), the AFOSR grant FA9550-17-1-0013, and the DARPA-AIRA grant HR00111990025. The work of Pengzhan Jin is partially supported by the Major Project on New Generation of Artificial Intelligence from the Ministry of Science and Technology of China (Grant No. 2018AAA010100).

\bibliographystyle{plain}
\bibliography{main}

\appendix

\section{Neural networks to approximate nonlinear operators}

We list in Table \ref{tab:notation} the main symbols and notations that are used throughout this paper.

\begin{table}[htbp]
\centering
\caption{Notations.}
\label{tab:notation}
\begin{tabular}{ll}
\toprule
$X$ & a Banach space with norm $\| \cdot \|_X$ \\
$\mathbb{R}^d$ & Euclidean space of dimension $d$ \\
$K$ & a compact set in a Banach space \\
$C(K)$ & Banach space of all continuous functions defined on $K$ with norm $\|f \|_{C(K)} = \max_{x \in K}|f(x)|$ \\
$V$ & a compact set in $C(K)$ \\
$u(x)$ & an input function/signal \\
$s(x)$ & an output function/signal \\
$f$ & a function or functional \\
$G$ & an operator \\
(TW) & all the Tauber-Wiener functions \\
$\sigma$ & an activation function \\
$\{x_1, x_2, \dots, x_m\}$ & $m$ sensor points \\
$n, p$ & neural network size hyperparameters in Theorems~\ref{thm:functional} and \ref{thm:operator} \\
$N$ & number of basis functions \\
$M$ & value of some upper bound \\
\bottomrule
\end{tabular}
\end{table}

Let $C(K)$ denote the Banach space of all continuous functions defined on a compact set $K \subset X$ with sup-norm $\|f \|_{C(K)} = \max_{x \in K}|f(x)|$, where $X$ is a Banach space. We first review the definition and sufficient condition of Tauber-Wiener (TW) functions~\cite{chen1995universal}, and the definition of continuous operator.

\begin{definition}[\textbf{TW}]
If a function $\sigma : \mathbb{R} \to \mathbb{R}$ (continuous or discontinuous) satisfies that all the linear combinations $\sum_{i=1}^N c_i\sigma(\lambda_ix+\theta_i)$, $\lambda_i \in \mathbb{R}$, $\theta_i \in \mathbb{R}$, $c_i \in \mathbb{R}$, $i=1,2,\dots,N$, are dense in every $C([a,b])$, then $\sigma$ is called a Tauber-Wiener (TW) function.
\end{definition}

\begin{theorem}[\textbf{Sufficient condition for TW}]
Suppose that $\sigma$ is a continuous function, and $\sigma \in S'(\mathbb{R})$ (tempered distributions), then $\sigma \in (TW)$, if and only if $\sigma$ is not a polynomial.
\end{theorem}

It is easy to verify that all the activation functions we used nowadays, such as sigmoid, $\tanh$ and ReLU, are TW functions.

\begin{definition}[\textbf{Continuity}]
Let $G$ be an operator between topological spaces $X$ and $Y$. We call $G$ continuous if for every $\epsilon > 0$, there exists a constant $\delta >0$ such that
$$\|G(x)-G(y)\|_Y < \epsilon$$
for all $x, y \in X$ satisfying $\|x-y\|_X < \delta$.
\end{definition}

We recall the following two main theorems of approximating nonlinear continuous functionals and operators due to Chen \& Chen~\cite{chen1995universal}.

\begin{theorem}[\textbf{Universal Approximation Theorem for Functional}]
\label{thm:functional}
Suppose that $\sigma \in (TW)$, $X$ is a Banach Space, $K \subset X$ is a compact set, $V$ is a compact set in $C(K)$, $f$ is a continuous functional defined on $V$, then for any $\epsilon > 0$, there are a positive integer $n$, $m$ points $x_1, \dots, x_m \in K$, and real constants $c_i$, $\theta_i$, $\xi_{ij}$, $i=1,\dots,n$, $j=1,\dots,m$, such that
$$\left|f(u) - \sum_{i=1}^n c_i \sigma\left(\sum_{j=1}^m \xi_{ij}u(x_j)+\theta_i\right)\right|<\epsilon$$
holds for all $u \in V$.
\end{theorem}

\begin{theorem}[\textbf{Universal Approximation Theorem for Operator}]
\label{thm:operator}
Suppose that $\sigma \in (TW)$, $X$ is a Banach Space, $K_1 \subset X$, $K_2 \subset \mathbb{R}^d$ are two compact sets in $X$ and $\mathbb{R}^d$, respectively, $V$ is a compact set in $C(K_1)$, $G$ is a nonlinear continuous operator, which maps $V$ into $C(K_2)$, then for any $\epsilon>0$, there are positive integers $n$, $p$, $m$, constants $c_i^k, \xi_{ij}^k, \theta_i^k, \zeta_k \in \mathbb{R}$, $w_k \in \mathbb{R}^d$, $x_j \in K_1$, $i=1,\dots,n$, $k=1,\dots,p$, $j=1,\dots,m$, such that
$$\left|G(u)(y) - \sum_{k=1}^p\sum_{i=1}^n c_i^k \sigma\left(\sum_{j=1}^m \xi_{ij}^ku(x_j)+\theta_i^k\right)\sigma(w_k \cdot y+\zeta_k) \right|<\epsilon$$
holds for all $u \in V$ and $y \in K_2$.
\end{theorem}

Both Theorems~\ref{thm:functional} and \ref{thm:operator} require the compactness of the function space. In fact, the function space like $C([0, 1])$ is ``too large'' for real applications, so it is suitable to consider a smaller space with compactness property. The necessary and sufficient conditions of the compactness are given by the following Arzelà–Ascoli Theorem.

\begin{definition}[\textbf{Uniform Boundedness}]
Let $V$ be a family of real-valued functions on the set $K$. We call $V$ uniformly bounded if there exists a constant $M$ such that
$$|f(x)| \leq M$$
for all $f \in V$ and all $x \in K$.
\end{definition}

\begin{definition}[\textbf{Equicontinuity}]
Let $V$ be a family of real-valued functions on the set $K$. We call $V$ equicontinuous if for every $\epsilon > 0$, there exists a $\delta > 0$ such that
$$|f(x)-f(y)| < \epsilon$$
for all $f \in V$ and all $x, y \in K$ satisfying $|x-y| < \delta$.
\end{definition}

\begin{theorem}[\textbf{Arzelà–Ascoli Theorem}]
Let $X$ be a Banach space, and $K \subset X$ be a compact set. A subset $V$ of $C(K)$ is pre-compact (has compact closure) if and only if
$V$ is uniformly bounded and equicontinuous.
\end{theorem}






\section{Number of sensors for identifying nonlinear dynamic systems}
\label{apd:m}

\begin{lemma}\label{lem:W_compact}
$W\coloneqq\bigcup_{i=1}^{\infty}W_{i}$ is compact.
\end{lemma}
\begin{proof}
At first, we prove that $W$ is pre-compact. For any $\varepsilon>0$, by (\ref{eq:max_bound}), there exists an $m_{0}$ such that
\begin{equation*}
    \|u-\mathcal{L}_{m}(u)\|_{C}<\frac{\varepsilon}{4},\quad\forall u\in V,\forall m>m_{0}.
\end{equation*}
Since $W_{m_{0}}$ is a compact set subject to equicontinuity, there exists a $\delta>0$ such that
\begin{equation*}
    |x-y|<\delta\Rightarrow|u(x)-u(y)|<\frac{\varepsilon}{2},\quad \forall u\in W_{m_{0}},\forall x,y\in [a,b].
\end{equation*}
Now for all $u\in W$ and all $x,y\in [a,b], |x-y|<\delta$, if $u\in W_{m_{0}}$, naturally we have $|u(x)-u(y)|<\frac{\varepsilon}{2}<\varepsilon$, otherwise, $u\in \bigcup_{i=m_{0}+1}^{\infty}U_{i}$. Suppose that $u=\mathcal{L}_{m}(v),m>m_{0},v\in V$, then there holds
\begin{equation*}
\begin{split}
    |u(x)-u(y)|= &|u(x)-v(x)+v(x)-v(y)+v(y)-u(y)| \\
    \leq &|\mathcal{L}_{m}(v)(x)-v(x)|+|v(x)-v(y)|+|\mathcal{L}_{m}(v)(y)-v(y)| \\
    \leq &2\|\mathcal{L}_{m}(v)-v\|_{C}+|v(x)-v(y)| \\
    < &2\cdot \frac{\varepsilon}{4}+\frac{\varepsilon}{2}=\varepsilon,
\end{split}
\end{equation*}
which shows the quicontinuity of $W$. In addition, it is obvious that $W$ is uniformly bounded, so that we know $W$ is pre-compact by applying the Arzelà–Ascoli Theorem.

Next we show that $W$ is close. Let $\{w_{i}\}_{i=1}^{\infty}\subset W$ be a sequence which converges to a $w_{0}\in C[a,b]$. If there exists an $m$ such that $\{w_{i}\}\subset W_{m}$, then $w_{0}\in W_{m}\subset W$. Otherwise, there is a subsequence $\{\mathcal{L}_{i_{n}}(v_{i_{n}})\}$ of $\{w_{i}\}$ such that $v_{i_{n}}\in V$ and $i_{n}\rightarrow\infty$ as $n\rightarrow\infty$. Then we have
\begin{equation*}
\begin{split}
    \|v_{i_{n}}-w_{0}\|_{C}=&\|v_{i_{n}}-\mathcal{L}_{i_{n}}(v_{i_{n}})+\mathcal{L}_{i_{n}}(v_{i_{n}})-w_{0}\|_{C} \\ 
    \leq&\|v_{i_{n}}-\mathcal{L}_{i_{n}}(v_{i_{n}})\|_{C}+\|\mathcal{L}_{i_{n}}(v_{i_{n}})-w_{0}\|_{C} \\
    \leq&\kappa(i_{n},V)+\|\mathcal{L}_{i_{n}}(v_{i_{n}})-w_{0}\|_{C},
\end{split}
\end{equation*}
which implies that $w_{0}=\lim_{n\rightarrow\infty}v_{i_{n}}\in V\subset W$.
\end{proof}
Next we show the proof of Theorem \ref{thm:m}.
\begin{proof}
For $u\in V$ and $u_{m}\in U_{m}$, according to the bound (\ref{eq:max_bound}) and the Lipschitz condition, we can derive that
\begin{eqnarray*}
\|(Gu)(d)-(Gu_m)(d)\|_{2} & \leq & c\int_a^d\|(Gu)(t)-(Gu_m)(t)\|_{2}dt+c\int_a^d|u(t)-u_m(t)|dt\\
&\leq&c\int_a^d\|(Gu)(t)-(Gu_m)(t)\|_{2}dt+c(b-a)\kappa(m,V).
\end{eqnarray*}
By Gronwall inequality, we have
\begin{equation*}
    \|(Gu)(d)-(Gu_m)(d))\|_{2} \leq c(b-a)\kappa(m,V)e^{c(b-a)}.
\end{equation*}

Define $S_m=\{(u(x_0),u(x_1),\cdots,u(x_m))\in\mathbb{R}^{m+1}|u\in V\}$ which is a compact set in $\mathbb{R}^{m+1}$, and there is a bijective mapping between $S_{m}$ and $U_{m}$. Furthermore, we define a vector-valued function on $S_m$ by 
\begin{equation*}
    \boldsymbol{\varphi}(u(x_0),u(x_1),\cdots,u(x_m))=(Gu_m)(d).
\end{equation*}
For any $\varepsilon>0$, make $m$ large enough so that $c(b-a)\kappa(m,V)e^{c(b-a)}<\varepsilon$. By universal approximation theorem of neural network for high-dimensional functions, there exist $\mathcal{W}_{1}\in \mathbb{R}^{n\times (m+1)},b_{1}\in \mathbb{R}^{m+1},\mathcal{W}_{2}\in \mathbb{R}^{K\times n},b_{2}\in \mathbb{R}^{K}$, such that
\begin{equation*}
    \|\boldsymbol{\varphi}(u(x_{0}),\cdots,u(x_{m}))-(\mathcal{W}_{2}\cdot\sigma(\mathcal{W}_{1}\cdot [u(x_{0})\quad\cdots\quad u(x_{m})]^{T}+b_{1})+b_{2})\|_{2}<\varepsilon-c(b-a)\kappa(m,V)e^{c(b-a)}.
\end{equation*}
Hence we have
\begin{equation*}
\begin{split}
    &\|(Gu)(d)-(\mathcal{W}_{2}\cdot\sigma(\mathcal{W}_{1}\cdot [u(x_{0})\quad\cdots\quad u(x_{m})]^{T}+b_{1})+b_{2})\|_{2} \\
    \leq&\|(Gu)(d)-(Gu_{m})(d)\|_{2}+\|(Gu_{m})(d)-(\mathcal{W}_{2}\cdot\sigma(\mathcal{W}_{1}\cdot [u(x_{0})\quad\cdots\quad u(x_{m})]^{T}+b_{1})+b_{2})\|_{2} \\
    <&c(b-a)\kappa(m,V)e^{c(b-a)}+\varepsilon-c(b-a)\kappa(m,V)e^{c(b-a)}=\varepsilon.
\end{split}
\end{equation*}

In summary, by choosing the value of $m$ so that it makes $c(b-a)\kappa(m,V)e^{c(b-a)}$ less than $\varepsilon$ is sufficient to achieve accuracy $\varepsilon$.
\end{proof}

\section{Gaussian random field with the radial-basis function kernel}
\label{apd:grf}
Suppose that $X(t) \sim  \mathcal{G}(0,\exp(-\frac{|x-y|^{2}}{l^{2}}))$. Then
$$X(t)=\sqrt{2}(\pi)^{\frac{1}{4}}\int_{\mathbb{R}^{+}}(l)^{\frac{1}{2}}\cos(\omega t)\exp(-\frac{l^{2}\omega^{2}}{8})dW(\omega)-\sqrt{2}(\pi)^{\frac{1}{4}}\int_{\mathbb{R}^{+}}(l)^{\frac{1}{2}}\sin(\omega t)\exp(-\frac{l^{2}\omega^{2}}{8})dB(\omega),$$
where $W$ and $B$ are independent standard Brownian motions \cite{zhang2017numerical}. Apply the change of variable $\lambda=l\omega$ and write $X(t)$ as
$$X(t)=\sqrt{2}(\pi)^{\frac{1}{4}}\int_{\mathbb{R}^{+}}\cos(\frac{\lambda}{l}t)\exp(-\frac{\lambda^{2}}{8})dW(\lambda)-\sqrt{2}(\pi)^{\frac{1}{4}}\int_{\mathbb{R}^{+}}\sin(\frac{\lambda}{l}t)\exp(-\frac{\lambda^{2}}{8})dB(\lambda).$$
Applying a linear interpolation $\Pi_{1}$ on the interval $[t_{i}, t_{i+1}]$, then
\begin{equation*}
    \begin{split}
        \mathbb{E}[(X(t)-\Pi_{1}X(t))^{2}]\ \ =&\ \ 2(\pi)^{\frac{1}{2}}\int_{\mathbb{R}^{+}}((I-\Pi_{1})\cos(\frac{\lambda}{l}t))^{2}\exp(-\frac{\lambda^{2}}{4})d\lambda \\
        &\ \ +2(\pi)^{\frac{1}{2}}\int_{\mathbb{R}^{+}}((I-\Pi_{1})\sin(\frac{\lambda}{l}t))^{2}\exp(-\frac{\lambda^{2}}{4})d\lambda \\
        \leq&\ \ (\pi)^{\frac{1}{2}}(t_{i+1}-t_{i})^{4}\int_{\mathbb{R}^{+}}(\frac{\lambda}{l})^{4}\exp(-\frac{\lambda^{2}}{4})d\lambda \\
        =&\ \ 24\pi\frac{(t_{i+1}-t_{i})^{4}}{l^{4}},
    \end{split}
\end{equation*}
where we recalled the error estimate of the linear interpolation on $[a,b]$ (by Taylor's expansion)
$$|(I-\Pi_{1})g(t)|=\frac{1}{2}(b-a)^{2}|f''(\xi)|,$$
where $\xi$ lies in between $a$ and $b$. Then by the Borel-Cantelli lemma, we have
$$|X(t)-\Pi_{1}X(t)|\leq C\frac{(t_{i+1}-t_{i})^{2-\epsilon}}{l^{2}},\quad\epsilon>0,$$
where $C$ is an absolute value of a Gaussian random variable with a finite variance. Therefore, taking a piecewise linear interpolation of $X(t)$ with $m$ points will lead to convergence with order $\mathcal{O}(\frac{1}{m^{2}l^{2}})$.

\end{document}